\pdfoutput=1

\documentclass[11pt]{article}

\usepackage{acl}

\usepackage{times}
\usepackage{latexsym}
\usepackage{multirow}
\usepackage{booktabs}
\usepackage{graphicx}
\usepackage{xspace}
\usepackage{amsmath}
\usepackage{dsfont}
\usepackage{scrextend}
\usepackage{subcaption}
\usepackage{listings}
\usepackage{color}
\usepackage{amsthm,amsfonts,amssymb,bm,stmaryrd,bbm}

\newcommand{\diffcse}{DiffCSE\xspace}
\newcommand{\la}{$_\texttt{large}$\xspace}
\newcommand{\ba}{$_\texttt{base}$\xspace}

\newcommand\tf[1]{\textbf{#1}}
\newcommand\ttt[1]{\texttt{#1}}

\newcommand\mr[1]{\mathrm{#1}}

\newcommand{\tableindent}{~~}
\newcommand\footnoteref[1]{\protected@xdef\@thefnmark{\ref{#1}}\@footnotemark}

\newcommand{\cls}{\ttt{[CLS]}}

\usepackage[T1]{fontenc}

\usepackage[utf8]{inputenc}

\usepackage{microtype}

\title{DiffCSE: Difference-based Contrastive Learning for Sentence Embeddings}

\author{Yung-Sung Chuang$^\dagger$ \quad Rumen Dangovski$^\dagger$ \quad Hongyin Luo$^\dagger$ \quad Yang Zhang$^\ddagger$ \quad Shiyu Chang$^\ast$ \\
\bf Marin Solja\v{c}i\'{c}$^\dagger$ \quad Shang-Wen Li$^\diamond$ \quad Wen-tau Yih$^\diamond$ \quad Yoon Kim$^{\dagger}$ \quad James Glass$^\dagger$ \\
  Massachusetts Institute of Technology$^\dagger$ \quad
  Meta AI$^\diamond$ \\
  MIT-IBM Watson AI Lab$^\ddagger$ \quad UC Santa Barbara$^\ast$ \\
  \texttt{yungsung@mit.edu} \\
    }
\begin{document}
\maketitle
\begin{abstract}
\vspace{-1mm}
We propose \diffcse, an unsupervised contrastive learning framework for learning sentence embeddings. \diffcse learns sentence embeddings that are sensitive to the difference between the original sentence and an edited sentence, where the edited sentence is obtained by stochastically masking out the original sentence and then sampling from a masked language model. We show that DiffSCE is an instance of equivariant contrastive learning \cite{dangovski2021equivariant}, which generalizes contrastive learning and learns representations that are insensitive to certain types of augmentations and sensitive to other ``harmful'' types of augmentations. Our experiments show that DiffCSE achieves state-of-the-art results among unsupervised sentence representation learning methods, outperforming unsupervised SimCSE\footnote{SimCSE has two settings: unsupervised and supervised. In this paper, we focus on the unsupervised setting. Unless otherwise stated, in this paper we use SimCSE to refer to unsupervised SimCSE.} by 2.3 absolute points on semantic textual similarity tasks.~\footnote{Pretrained models and code are available at {\url{https://github.com/voidism/DiffCSE}}.}
\vspace{-2mm}
\end{abstract}
\vspace{-1mm}
\section{Introduction}
\vspace{-1mm}
Learning ``universal'' sentence representations that capture rich semantic information and are at the same time performant across a wide range of downstream NLP tasks without task-specific finetuning is an important open issue in the field~\cite{conneau2017supervised, cer2018universal, kiros2015skip-thought, logeswaran2018an-quick-thought, giorgi2020declutr, yan2021consert, gao2021simcse}. Recent work has shown that finetuning pretrained language models with  \emph{contrastive learning} makes it possible to learn good sentence embeddings without  any labeled data~\cite{giorgi2020declutr, yan2021consert, gao2021simcse}. Contrastive learning uses multiple augmentations on a single datum to construct positive pairs whose representations are trained to be more similar to one another than negative pairs. While different data augmentations (random cropping, color jitter, rotations, etc.) have been found to be crucial for pretraining vision models~\cite{chen2020simple}, such augmentations have generally been unsuccessful when applied to contrastive learning of sentence embeddings. Indeed,  \citet{gao2021simcse} find that constructing positive pairs via a simple dropout-based augmentation works much better than more complex augmentations such as word deletions or replacements based on synonyms or masked language models. This is perhaps unsurprising in hindsight; while the training objective in contrastive learning encourages representations to be \emph{invariant} to augmentation transformations, direct augmentations on the input (e.g., deletion, replacement) often change the meaning of the sentence. That is, ideal sentence embeddings should \emph{not} be invariant to such transformations.

\begin{figure*}[!h]
    \centering
        \vspace{-2mm}
    \includegraphics[width=0.8\linewidth]{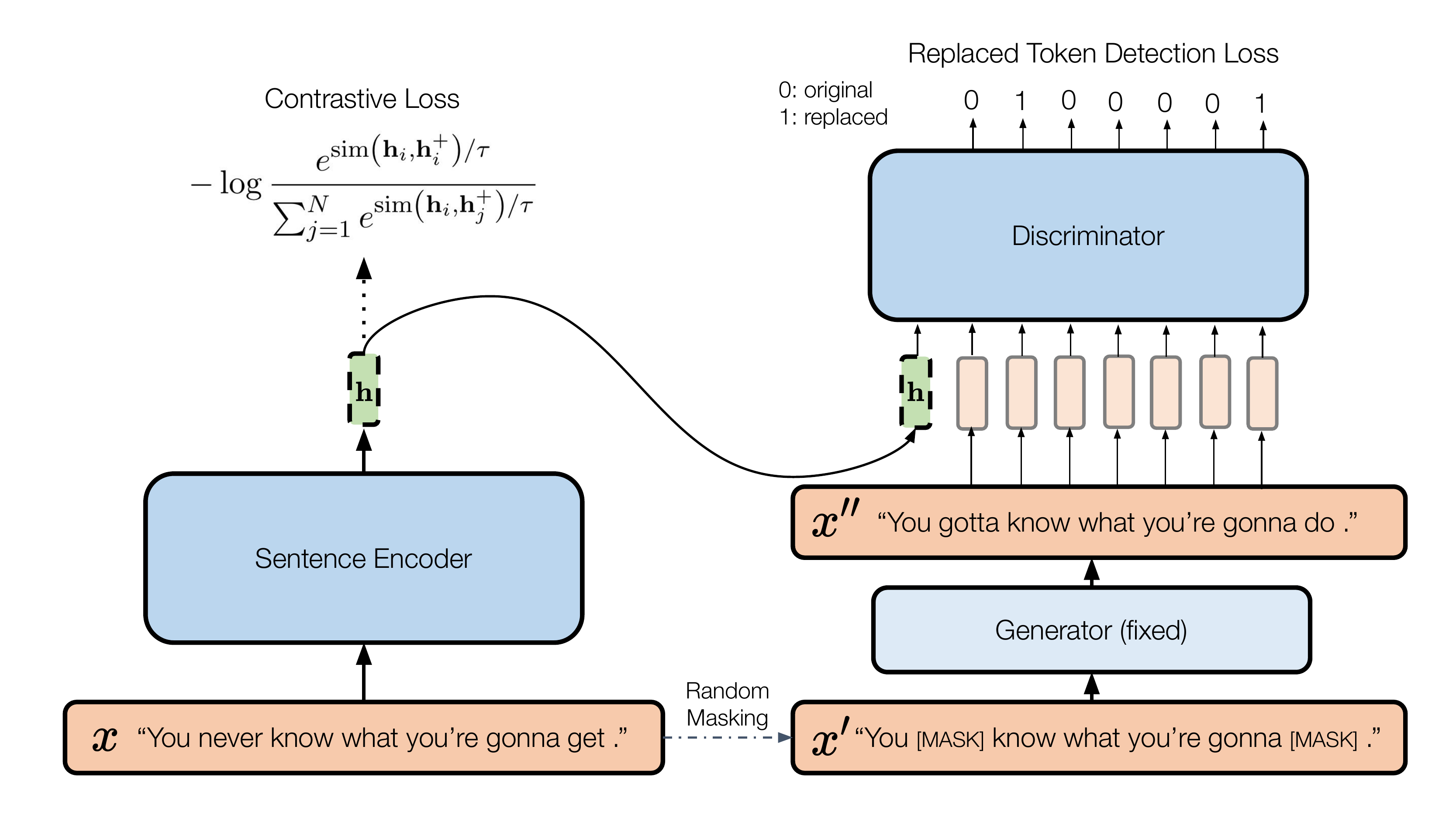}
    \caption{Illustration of DiffCSE. On the left-hand side is a standard SimCSE model trained with regular contrastive loss on dropout transformations. On the right hand side is a conditional difference prediction model which takes the sentence vector $\mathbf{h}$ as input and predict the difference between $x$ and $x^{\prime\prime}$. During testing we discard the discriminator and only use $\mathbf{h}$ as the sentence embedding.}
    \label{fig:diffcse}
    \vspace{-2mm}
\end{figure*}

We propose to learn sentence representations that are \emph{aware} of, but not necessarily invariant to, such direct surface-level augmentations. This is an instance of \emph{equivariant} contrastive learning \cite{dangovski2021equivariant}, which improves vision representation learning by using a contrastive loss on  \emph{insensitive} image transformations (e.g., grayscale) and a prediction loss on \emph{sensitive} image transformations (e.g., rotations). We operationalize equivariant contrastive learning on sentences by using dropout-based augmentation as the insensitive transformation (as in SimCSE \cite{gao2021simcse}) and MLM-based word replacement as the sensitive transformation. This results in an additional cross-entropy loss based on the \emph{difference} between the original and the transformed sentence.

We conduct experiments on 7 semantic textual similarity tasks (STS) and 7 transfer tasks from SentEval \cite{conneau-kiela-2018-senteval} and find that this difference-based learning greatly improves over standard contrastive learning. Our DiffCSE approach can achieve around 2.3\% absolute improvement on STS datasets over SimCSE, the previous state-of-the-art model. We also conduct a set of ablation studies to justify our designed architecture. Qualitative study and analysis are also included to look into the embedding space of DiffCSE.

\section{Background and Related Work}
\subsection{Learning Sentence Embeddings}
Learning universal sentence embeddings has been studied extensively in prior work, including unsupervised approaches such as Skip-Thought~\cite{kiros2015skip-thought}, Quick-Thought~\cite{logeswaran2018an-quick-thought} and FastSent~\cite{hill2016learning}, or supervised methods such as InferSent~\cite{conneau2017supervised}, Universal Sentence Encoder~\cite{cer2018universal} and Sentence-BERT~\cite{reimers2019sentence}. Recently, researchers have focused on (unsupervised) contrastive learning approaches such as SimCLR~\cite{chen2020simple} to learn sentence embeddings.
SimCLR~\cite{chen2020simple} learns image representations by creating semantically close augmentations for the same images and then pulling these representations to be closer than representations of random negative examples. The same framework can be adapted to learning sentence embeddings by designing good augmentation methods for natural language.
ConSERT~\cite{yan2021consert} uses a combination of four data augmentation strategies: adversarial attack, token shuffling, cut-off, and dropout.
DeCLUTR~\cite{giorgi2020declutr} uses overlapped spans as positive examples and distant spans as negative examples for learning contrastive span representations.
Finally, SimCSE~\cite{gao2021simcse} proposes an extremely simple augmentation strategy by just switching dropout masks. While simple, sentence embeddings learned in this manner have been shown to be better than other more complicated augmentation methods.

\subsection{Equivariant Contrastive Learning}
\label{sec:ecl}
\diffcse is inspired by  a recent generalization of contrastive learning in computer vision (CV) called equivariant contrastive learning~\citep{dangovski2021equivariant}. We now explain how this CV  technique can be adapted to natural language.

Understanding the role of input transformations is crucial for successful contrastive learning. Past empirical studies have revealed useful transformations for contrastive learning, such as random resized cropping and color jitter for computer vision \cite{chen2020simple} and dropout for NLP \cite{gao2021simcse}. Contrastive learning encourages representations to be insensitive to these transformations, i.e. the encoder is trained to be invariant to a set of manually chosen transformations. 
The above studies in CV and NLP have also revealed transformations that are \emph{harmful} for contrastive learning. For example,~\citet{chen2020simple} showed that making the representations insensitive to rotations decreases the ImageNet linear probe accuracy, and~\citet{gao2021simcse} showed that using an MLM to replace 15\% of the words drastically reduces performance on STS-B. While previous works simply omit these transformations from contrastive pre-training, here we argue that we should still make use of these transformations by learning representations that are  \emph{sensitive} (but not necessarily invariant) to such transformations.

The notion of (in)sensitivity can be captured by the more general property of equivariance in mathematics. Let $T$ be a transformation from a group $G$ and let $T(x)$ denote the transformation of a sentence $x$. Equivariance is the property that there is an induced group transformation $T'$ on the output features~\citep{dangovski2021equivariant}:
\vspace{-5pt}
\[
f(T(x))=T'(f(x)).
\vspace{-5pt}
\]
In the special case of contrastive learning, $T'$'s target is the identity transformation, and we say that $f$ is trained to be ``invariant to $T$.'' However, invariance is just a trivial case of equivariance, and we can design training objectives where $T'$ is not the identity for some transformations (such as MLM), while it is the identity for others (such as dropout). \citet{dangovski2021equivariant}~show that generalizing contrastive learning to equivariance in this way improves the semantic quality of features in CV, and here we show that the complementary nature of invariance and equivariance extends to the NLP domain. The key observation is that the encoder should be equivariant to MLM-based augmentation instead of being invariant. 
We can operationalize this by using a conditional discriminator that combines the sentence representation with an edited sentence, and then predicts the \emph{difference} between the original and edited sentences. This is essentially a conditional version of the ELECTRA model~\citep{clark2020electra}, which makes the encoder equivariant to MLM by using a binary discriminator which detects whether a token is from the original sentence or from a generator. We hypothesize that conditioning the ELECTRA model with the representation from our sentence encoder is a useful objective for encouraging $f$ to be ``equivariant to MLM.''

To the best of our knowledge, we are the first to observe and highlight the above parallel between CV and NLP. In particular, we show that equivariant contrastive learning extends beyond CV, and that it works for transformations even without algebraic structures, such as diff operations on sentences. Further, insofar as the canonical set of useful transformations is less established in NLP than is in CV, DiffCSE can serve as a diagnostic tool for NLP researchers to discover useful transformations.

\section{Difference-based Contrastive Learning}
\label{sec:method}

Our approach is straightforward and can be seen as combining the standard contrastive learning objective from SimCSE (Figure~\ref{fig:diffcse}, left) with a \emph{difference prediction} objective which conditions on the sentence embedding (Figure~\ref{fig:diffcse}, right).

Given an unlabeled input sentence $x$, SimCSE creates a positive example $x^{+}$ for it by applying different dropout masks. By using the {BERT\ba} encoder $f$, we can obtain the sentence embedding $\mathbf{h}=f\left(x\right)$ for $x$ (see section~\ref{sec:exp} for how $\mathbf{h}$ is obtained). The training objective for SimCSE is:
\vspace{-5pt}
\begin{align*}
    \mathcal{L}_{\text {contrast }} = -\log \frac{e^{\operatorname{sim}\left(\mathbf{h}_{i}, \mathbf{h}_{i}^{+}\right) / \tau}}{\sum_{j=1}^{N}e^{\operatorname{sim}\left(\mathbf{h}_{i}, \mathbf{h}_{j}^{+}\right) / \tau}},
    \vspace{-5pt}
\end{align*}
where $N$ is the batch size for the input batch $\left\{x_{i}\right\}_{i=1}^{N}$ as we are using in-batch negative examples, $\operatorname{sim}(\cdot, \cdot)$ is the cosine similarity function, and $\tau$ is a temperature hyperparameter.

On the right-hand side of Figure~\ref{fig:diffcse} is a conditional version of the difference prediction objective used in ELECTRA \cite{clark2020electra}, which contains a generator and a discriminator. Given a sentence of length $T$, $x=[x_{(1)}, x_{(2)}, ..., x_{(T)}]$, we first apply a random mask $m=[m_{(1)}, m_{(2)}, ..., m_{(T)}], m_{(t)} \in [0, 1]$ on $x$ to obtain $x^\prime = m \cdot x$. We use another pretrained MLM as the generator $G$ to perform masked language modeling to recover randomly masked tokens in $x^\prime$ to obtain the edited sentence $x^{\prime\prime}=G(x^\prime)$. Then, we use a discriminator $D$ to perform the Replaced Token Detection (RTD) task. For each token in the sentence, the model needs to predict whether it has been replaced or not. The cross-entropy loss for a single sentence $x$ is:
\vspace{-5pt}
\begin{align*}
    \mathcal{L}^x_{\text {RTD }}&=\sum_{t=1}^{T}\biggl(-\mathds{1}\left(x_{(t)}^{\prime\prime}=x_{(t)}\right) \log D\left(x^{\prime\prime}, \mathbf{h}, t\right) \\
    &-\mathds{1}\left(x_{(t)}^{\prime\prime} \neq x_{(t)}\right) \log \left(1-D\left(x^{\prime\prime}, \mathbf{h}, t\right)\right)\biggr)
    \vspace{-5pt}
\end{align*}
And the training objective for a batch is $\mathcal{L}_{\text {RTD }}= \sum_{i=1}^{N}\mathcal{L}^{x_i}_{\text {RTD }}$.
Finally we optimize these two losses together with a weighting coefficient $\lambda$:
\vspace{-5pt}
\begin{align*}
    \mathcal{L}&= \mathcal{L}_{\text {contrast }} + \lambda \cdot \mathcal{L}_{\text {RTD }}
    \vspace{-5pt}
\end{align*}
The difference between our model and ELECTRA is that our discriminator $D$ is \emph{conditional}, so it can use the information of $x$ compressed in a fixed-dimension vector $\mathbf{h} = f\left(x\right)$. The gradient of $D$ can be backward-propagated into $f$ through $\mathbf{h}$. By doing so, $f$ will be encouraged to make $\mathbf{h}$ informative enough to cover the full meaning of $x$, so that $D$ can distinguish the tiny difference between $x$ and $x^{\prime\prime}$. This approach essentially makes the conditional discriminator perform a ``diff operation'', hence the name DiffCSE.

When we train our \diffcse model, we fix the generator $G$, and only the sentence encoder $f$ and the discriminator $D$ are optimized. After training, we discard $D$ and only use $f$ (which remains fixed) to extract sentence embeddings to evaluate on the downstream tasks.

\begin{table*}[th!]
    \begin{center}
    \centering
    \small
    \begin{tabular}{lcccccccc}
    \toprule
       \tf{Model} & \tf{STS12} & \tf{STS13} & \tf{STS14} & \tf{STS15} & \tf{STS16} & \tf{STS-B} & \tf{SICK-R} & \tf{Avg.} \\
    \midrule
        GloVe embeddings (avg.)$^\clubsuit$ & 55.14 & 70.66 & 59.73 & 68.25 & 63.66 & 58.02 & 53.76 & 61.32 \\
        BERT\ba~(first-last avg.)$^\diamondsuit$ & 39.70&	59.38&	49.67&	66.03&	66.19&	53.87&	62.06&	56.70\\
        BERT\ba-flow$^\diamondsuit$ & 58.40&	67.10&	60.85&	75.16&	71.22&	68.66&	64.47&	66.55 \\ 
        BERT\ba-whitening$^\diamondsuit$ & 57.83& 66.90 & 60.90 & 75.08& 71.31& 68.24& 63.73& 66.28\\ 
        IS-BERT\ba$^\heartsuit$ & 56.77 & 69.24 & 61.21 & 75.23 & 70.16 & 69.21 & 64.25 & 66.58 \\
        CMLM-BERT\ba$^\spadesuit$ {\scriptsize(1TB data)} & 58.20 & 61.07 & 61.67 & 73.32 & 74.88 & 76.60 & 64.80 & 67.22 \\
        CT-BERT\ba$^\diamondsuit$ & 61.63 & 76.80 & 68.47 & 77.50 & 76.48 & 74.31 & 69.19 &72.05 \\
        SG-OPT-BERT\ba$^\dagger$ & 66.84 & 80.13 & 71.23 & 81.56 & 77.17 & 77.23 & 68.16 & 74.62 \\
        SimCSE-BERT\ba$^\diamondsuit$ &  68.40 & 82.41  & 74.38 & 80.91 & 78.56 & 76.85 & \bf 72.23 & 76.25 \\
        $*$ SimCSE-BERT\ba {\scriptsize(reproduce)} & 70.82 & 82.24 & 73.25 & 81.38 & 77.06 & 77.24 & 71.16 & 76.16 \\
        $*$ DiffCSE-BERT\ba & \bf 72.28 & \bf 84.43 & \bf 76.47 & \bf 83.90 & \bf 80.54 & \bf 80.59 & 71.23 & \bf 78.49 \\
        \midrule
        RoBERTa\ba~(first-last avg.)$^\diamondsuit$ & 40.88&	58.74&	49.07&	65.63&	61.48&	58.55&	61.63&	56.57\\
        RoBERTa\ba-whitening$^\diamondsuit$ & 46.99& 63.24&	57.23&	71.36&	68.99&	61.36&	62.91& 61.73\\ 
        DeCLUTR-RoBERTa\ba$^\diamondsuit$ & 52.41 & 75.19& 65.52 & 77.12 & 78.63 & 72.41 &  68.62  & 69.99\\
        SimCSE-RoBERTa\ba$^\diamondsuit$ &  \bf 70.16  & 81.77 	& 73.24 	& 81.36 	& 80.65 	& 80.22 	&	68.56 &  76.57 \\
        $*$ SimCSE-RoBERTa\ba {\scriptsize(reproduce)} &  68.60 & 81.36 & 73.16 & 81.61 & 80.76 & 80.58 & 68.83 & 76.41 \\
        $*$ DiffCSE-RoBERTa\ba & 70.05 & \bf 83.43 & \bf 75.49 & \bf 82.81 & \bf 82.12 & \bf 82.38 & \bf 71.19 & \bf 78.21 \\
    \bottomrule
    \end{tabular}
    \end{center}
\vspace{-3.5mm}
    \caption{
        The performance on STS tasks (Spearman's correlation) for different sentence embedding models.
        $\clubsuit$: results from \citet{reimers2019sentence};
        $\heartsuit$: results from \citet{zhang2020unsupervised};
        $\diamondsuit$: results from \citet{gao2021simcse};
        $\spadesuit$: results from \citet{yang2020universal};
        $\dagger$: results from \citet{kim2021self};
        $*$: results from our experiments.
    }
    \label{tab:main_sts}
\vspace{-2mm}
\end{table*}

\section{Experiments}
\label{sec:exp}
\subsection{Setup}
In our experiment, we follow the setting of unsupervised SimCSE~\cite{gao2021simcse} and build our model based on their PyTorch implementation.\footnote{\label{github}\scriptsize \url{https://github.com/princeton-nlp/SimCSE}} We also use the checkpoints of BERT~\cite{devlin2019bert} and RoBERTa~\cite{liu2019roberta} as the initialization of our sentence encoder $f$. We add an MLP layer with Batch Normalization~\cite{ioffe2015batch} (BatchNorm) on top of the \texttt{[CLS]} representation as the sentence embedding. We will compare the model with/without BatchNorm in section~\ref{sec:ablation}. For the discriminator $D$, we use the same model as the sentence encoder $f$ (BERT/RoBERTa). For the generator $G$, we use the smaller DistilBERT and DistilRoBERTa~\cite{sanh2019distilbert} for efficiency. Note that the generator is fixed during training unlike the ELECTRA paper~\cite{clark2020electra}. We will compare the results of using different size model for the generator in section~\ref{sec:ablation}. More training details are shown in Appendix~\ref{sec:appendix}.

\subsection{Data}
For unsupervised pretraining, we use the same $10^6$ randomly sampled sentences from English Wikipedia that are provided by the source code of SimCSE.\footref{github}
We evaluate our model on 7 semantic textual similarity (STS) and 7 transfer tasks in SentEval.\footnote{\scriptsize\url{https://github.com/facebookresearch/SentEval}} STS tasks includes STS 2012--2016~\cite{agirre-etal-2016-semeval},
STS Benchmark~\cite{cer-etal-2017-semeval} and
SICK-Relatedness~\cite{marelli2014sick}.
All the STS experiments are fully unsupervised, which means no STS training datasets are used and all embeddings are fixed once they are trained. 
The transfer tasks are various sentence classification tasks, including MR~\cite{pang2005seeing}, CR~\cite{hu2004mining}, SUBJ~\cite{pang2004sentimental}, MPQA~\cite{wiebe2005annotating}, SST-2~\cite{socher2013recursive}, TREC~\cite{voorhees2000building} and MRPC~\cite{dolan2005automatically}.
In these transfer tasks, we will use a logistic regression classifier trained on top of the frozen sentence embeddings, following the standard setup \cite{conneau-kiela-2018-senteval}.

\subsection{Results}

\paragraph{Baselines} We compare our model with many strong unsupervised baselines including SimCSE~\cite{gao2021simcse}, IS-BERT~\cite{zhang2020unsupervised}, CMLM~\cite{yang2020universal}, DeCLUTR~\cite{giorgi2020declutr}, CT-BERT~\cite{carlsson2021semantic}, SG-OPT~\cite{kim2021self} and some post-processing methods like BERT-flow~\cite{li-etal-2020-sentence} and BERT-whitening~\cite{su2021whitening} along with some naive baselines like averaged GloVe embeddings~\cite{pennington2014glove} and averaged first and last layer BERT embeddings.

\paragraph{Semantic Textual Similarity (STS)}
We show the results of STS tasks in Table~\ref{tab:main_sts} including BERT\ba (upper part) and RoBERTa\ba (lower part). We also reproduce the previous state-of-the-art SimCSE~\cite{gao2021simcse}. \diffcse-BERT\ba can significantly outperform SimCSE-BERT\ba and raise the averaged Spearman’s correlation from 76.25\% to 78.49\%. 
For the RoBERTa model, \diffcse-RoBERTa\ba can also improve upon SimCSE-RoBERTa\ba from 76.57\% to 77.80\%.

\paragraph{Transfer Tasks}
We show the results of transfer tasks in Table~\ref{tab:main_transfer}. Compared with SimCSE-BERT\ba, DiffCSE-BERT\ba can improve the averaged scores from 85.56\% to 86.86\%. When applying it to the RoBERTa model, DiffCSE-RoBERTa\ba also improves upon SimCSE-RoBERTa\ba from 84.84\% to 87.04\%.
Note that the CMLM-BERT\ba~\cite{yang2020universal} can achieve even better performance than DiffCSE. However, they use 1TB of the training data from Common Crawl dumps while our model only use 115MB of the Wikipedia data for pretraining. We put their scores in Table~\ref{tab:main_transfer} for reference.
In SimCSE, the authors propose to use MLM as an auxiliary task for the sentence encoder to further boost the performance of transfer tasks. Compared with the results of SimCSE with MLM, DiffCSE still can have a little improvement around 0.2\%.

\begin{table*}[ht]
    \begin{center}
    \centering
    \small

    \begin{tabular}{lcccccccc}
    \toprule
       \tf{Model} & \tf{MR} & \tf{CR} & \tf{SUBJ} & \tf{MPQA} & \tf{SST} & \tf{TREC} & \tf{MRPC} & \tf{Avg.}\\
    \midrule
        GloVe embeddings (avg.)$^\clubsuit$ & 77.25&    78.30&  91.17&  87.85&  80.18&  83.00& 72.87 & 81.52\\
        Skip-thought$^\heartsuit$ &  76.50& 80.10&  93.60&  87.10&  82.00&  92.20&  73.00& 83.50  \\
        \midrule
        Avg. BERT embeddings$^\clubsuit$ & 78.66 & 86.25 & 94.37 & 88.66 & 84.40 & \tf{92.80} & 69.54 & 84.94 \\
        BERT-\cls embedding$^\clubsuit$ & 78.68 & 84.85 & 94.21 & 88.23 & 84.13 & 91.40 & 71.13 & 84.66 \\
        IS-BERT\ba$^\heartsuit$ & 81.09 & 87.18 & 94.96 & 88.75 & 85.96 & 88.64 & 74.24 & 85.83 \\
        SimCSE-BERT\ba$^\diamondsuit$ & 81.18&	86.46&	94.45&	88.88&	85.50&	89.80&	74.43&	85.81\\
        \tableindent w/ MLM & 82.92&	87.23&	95.71&	88.73&	86.81&	87.01&	78.07& 86.64\\
        $*$ DiffCSE-BERT\ba & \bf 82.69 & \bf 87.23 & \bf 95.23 & \bf 89.28 & \bf 86.60 & 90.40 & \bf 76.58 & \bf 86.86\\
        \midrule
        CMLM-BERT\ba {\scriptsize(1TB data)} & 83.60 & 89.90 & 96.20 & 89.30 & 88.50 & 91.00 & 69.70 & 86.89 \\
        \midrule
        SimCSE-RoBERTa\ba$^\diamondsuit$ & 81.04 &	87.74 &	93.28 &	86.94	&86.60	&84.60	&73.68 &84.84 \\
        \tableindent w/ MLM & \tf{83.37}& 	87.76& 	\tf{95.05}	& 87.16	& \tf{89.02}	& \tf{90.80}	& 75.13 & 86.90\\
        $*$ DiffCSE-RoBERTa\ba & 82.82 & \bf 88.61 & 94.32 & \bf 87.71 & 88.63 & 90.40 & \bf 76.81 & \bf 87.04 \\
    \bottomrule
    \end{tabular}
    \end{center}
\vspace{-3.5mm}
    \caption{
        Transfer task results of different sentence embedding models (measured as accuracy). $\clubsuit$: results from \citet{reimers2019sentence};
        $\heartsuit$: results from \citet{zhang2020unsupervised};
        $\diamondsuit$: results from \citet{gao2021simcse}.
    }
    \vspace{-3mm}
    \label{tab:main_transfer}
\end{table*}

\section{Ablation Studies}
\vspace{-2mm}
\label{sec:ablation}
In the following sections, we perform an extensive series of ablation studies that support our model design. We use BERT\ba model to evaluate on the development set of STS-B and transfer tasks.
\vspace{-2mm}
\paragraph{Removing Contrastive Loss}
In our model, both the contrastive loss and the RTD loss are crucial because they maintain what should be sensitive and what should be insensitive respectively. If we remove the RTD loss, the model becomes a SimCSE model; if we remove the contrastive loss, the performance of STS-B drops significantly by 30\%, while the average score of transfer tasks also drops by 2\% (see Table~\ref{tab:ablation}). This result shows that it is important to have insensitive and sensitive attributes that exist together in the representation space.

\paragraph{Next Sentence vs. Same Sentence}
Some methods for unsupervised sentence embeddings like Quick-Thoughts~\cite{logeswaran2018an-quick-thought} and CMLM~\cite{yang2020universal} predict the next sentence as the training objective. We also experiment with a variant of DiffCSE by conditioning the ELECTRA loss based on the next sentence. Note that this kind of model is not doing a ``diff operation'' between two similar sentences, and is not an instance of equivariant contrastive learning.
As shown in Table~\ref{tab:ablation} (use next sent. for $x^\prime$), the score of STS-B decreases significantly compared to DiffCSE while transfer performance remains similar. We also tried using the same sentence and the next sentence at the same time for conditioning the ELECTRA objective (use same+next sent. for $x^\prime$), and did not observe improvements.

\begin{table}[t!]
    \begin{center}
    \centering
    \small
    \begin{tabular}{lcc}
    \toprule
         & \tf{STS-B} & \tf{Avg. transfer} \\
    \midrule
        SimCSE & 81.47 & 83.91 \\
    \midrule
        DiffCSE & \bf 84.56 & \bf 85.95\\
        ~~w/o contrastive loss & 54.48 & 83.46 \\
        ~~use next sent. for $x^{\prime}$ & 82.91 & 85.83 \\
        ~~use same+next sent. for $x^{\prime}$ & 83.41 & 85.82 \\
    \midrule
        Conditional MLM & & \\
        ~~for same sent. & 83.08 & 84.43 \\
        ~~for next sent. & 75.82 & 85.68 \\
        ~~for same+next sent. & 82.88 & 84.82 \\
    \midrule
        Conditional Corrective LM & 79.79 & 85.30 \\
    \bottomrule
    \end{tabular}
    \end{center}
\vspace{-3mm}
    \caption{
        Development set results of STS-B and transfer tasks for DiffCSE model variants, where we vary the objective and the use of same or next sentence.
    }
    \label{tab:ablation}
\vspace{-2mm}
\end{table}

\begin{table}[ht]
\centering
\small
\setlength\tabcolsep{3pt}
\begin{tabular}{lcc}
\toprule
\bf Augmentation               & \bf STS-B & \multicolumn{1}{c}{\bf Avg. transfer} \\
\midrule

MLM 15\%                   & \bf 84.48 & 85.95                                    \\
randomly insert 15\%         & 82.20 & 85.96                                    \\
randomly delete 15\%         & 82.59 & \bf 85.97                                    \\
combining all              & 82.80 & 85.92 \\
\bottomrule

\end{tabular}
\vspace{-2mm}
\caption{Development set results of STS-B and transfer tasks with different augmentation methods for learning equivariance.}
\label{tab:augments}
\vspace{-4mm}
\end{table}

\paragraph{Other Conditional Pretraining Tasks}
Instead of a conditional binary difference prediction loss, we can also consider other conditional pretraining tasks such as a conditional MLM objective proposed by \citet{yang2020universal}, or corrective language modeling,\footnote{This task is similar to ELECTRA. However, instead of a binary classifier for replaced token detection, corrective LM uses a vocabulary-size classifier with the copy mechanism to recover the replaced tokens.} proposed by COCO-LM~\cite{meng2021coco}. We experiment with these objectives instead of the difference prediction objective in Table~\ref{tab:ablation}. We observe that conditional MLM on the same sentence does not improve the performance either on STS-B or transfer tasks compared with DiffCSE.
Conditional MLM on the next sentence performs even worse for STS-B, but slightly better than using the same sentence on transfer tasks. Using both the same and the next sentence also does not improve the performance compared with DiffCSE.
For the corrective LM objective, the performance of STS-B decreases significantly compared with DiffCSE.

\begin{table}[h!]
    \begin{center}
    \centering
    \small
    \begin{tabular}{lcc}
    \toprule
         & \tf{STS-B} & \tf{Avg. transfer} \\
    \midrule
        DiffCSE & \\
        ~~w/ BatchNorm & \bf 84.56 & \bf 85.95 \\
        ~~w/o BatchNorm & 83.23 & 85.24 \\
    \midrule
        SimCSE & \\
        ~~w/ BatchNorm & 82.22 & 85.66 \\
        ~~w/o BatchNorm & 81.47 & 83.91 \\
    \bottomrule
    \end{tabular}
    \end{center}
\vspace{-2mm}
    \caption{
        Development set results of STS-B and transfer tasks for DiffCSE and SimCSE with and without BatchNorm. 
    }
    \label{tab:bn}
\vspace{-4mm}
\end{table}

\paragraph{Augmentation Methods: Insert/Delete/Replace}

In DiffCSE, we use MLM token replacement as the equivariant augmentation. It is possible to use other methods like random insertion or deletion instead of replacement.\footnote{Edit distance operators include \emph{insert, delete} and \emph{replace}.} For insertion, we choose to randomly insert mask tokens to the sentence, and then use a generator to convert mask tokens into real tokens. The number of inserted masked tokens is 15\% of the sentence length. The task is to predict whether a token is an inserted token or the original token. For deletion, we randomly delete 15\% tokens in the sentence, and the task  is to predict for each token whether a token preceding it has been deleted or not. The results are shown in Table~\ref{tab:augments}. We can see that using either insertion or deletion achieves a slightly worse STS-B performance than using MLM replacement. For transfer tasks, their results are similar. Finally, we find that combining all three augmentations in the training process does not improve the MLM replacement strategy.

\paragraph{Pooler Choice}

In SimCSE, the authors use the pooler in BERT’s original implementation (one linear layer with tanh activation function) as the final layer to extract features for computing contrastive loss. In our implementation (see details in Appendix~\ref{sec:appendix}), we find that it is better to use a two-layer pooler with Batch Normalization (BatchNorm)~\cite{ioffe2015batch}, which is commonly used in contrastive learning framework in computer vision~\cite{chen2020simple, grill2020byol, chen2021exploring, hua2021feature}. We show the ablation results in Table~\ref{tab:bn}. We can observe that adding BatchNorm is beneficial for either DiffCSE or SimCSE to get better performance on STS-B and transfer tasks.

\paragraph{Size of the Generator}

In our DiffCSE model, the generator can be in different model size from BERT\la, BERT\ba~\cite{devlin2019bert}, DistilBERT\ba~\cite{sanh2019distilbert}, BERT$_\texttt{medium}$, BERT$_\texttt{small}$, BERT$_\texttt{mini}$, BERT$_\texttt{tiny}$~\cite{turc2019well}. Their exact sizes are shown in Table~\ref{tab:gsize} (L: number of layers, H: hidden dimension). Notice that although DistilBERT\ba has only half the number of layers of BERT, it can retain 97\% of BERT's performance due to knowledge distillation.

We show our results in Table~\ref{tab:gsize}, we can see the performance of transfer tasks does not change much with different generators. However, the score of STS-B decreases as we switch from BERT-medium to BERT-tiny. This finding is not the same as ELECTRA, which works best with generators 1/4-1/2 the size of the discriminator. Because our discriminator is conditional on sentence vectors, it will be easier for the discriminator to perform the RTD task. As a result, using stronger generators (BERT\ba, DistilBERT\ba) to increase the difficulty of RTD would help the discriminator learn better. However, when using a large model like BERT\la, it may be a too-challenging task for the discriminator. In our experiment, using DistilBERT\ba, which has the ability close to but slightly worse than BERT\ba, gives us the best performance.

\begin{table}[t!]
    \begin{center}
    \centering
    \small
    \begin{tabular}{lcc}
    \toprule
         & \tf{STS-B} & \tf{Avg. transfer} \\
    \midrule
        SimCSE & 81.47 & 83.91 \\
    \midrule
        DiffCSE w/ generator:& & \\
        ~~BERT$_\texttt{large}$ (L=24, H=1024) & 82.93 & 85.88 \\
        ~~BERT$_\texttt{base}$ (L=12, H=768) & 83.63 & 85.85 \\
        ~~DistilBERT$_\texttt{base}$ (L=6, H=768) & \bf 84.56 & \bf 85.95 \\
    \midrule
        ~~BERT$_\texttt{medium}$ (L=8, H=512) & 82.25 & 85.80 \\
        ~~BERT$_\texttt{small}$ (L=4, H=512)  & 82.64 & 85.66 \\
        ~~BERT$_\texttt{mini}$ (L=4, H=256) & 82.12 & 85.90 \\
        ~~BERT$_\texttt{tiny}$ (L=2, H=128) & 81.40 & 85.23 \\
    \bottomrule
    \end{tabular}
    \end{center}
\vspace{-3mm}
    \caption{
        Development set results of STS-B and transfer tasks with different generators. 
    }
    \label{tab:gsize}
    \vspace{-1mm}
\end{table}

\paragraph{Masking Ratio}

In our conditional ELECTRA task, we can mask the original sentence in different ratios for the generator to produce MLM-based augmentations. A higher masking ratio will make more perturbations to the sentence. Our empirical result in Table~\ref{tab:mask} shows that the difference between difference masking ratios is small (in 15\%-40\% ), and a masking ratio of around 30\% can give us the best performance.

\begin{table}[t!]
    \begin{center}
    \centering
    \small
    \begin{tabular}{lp{0.6cm}p{0.6cm}p{0.6cm}p{0.6cm}p{0.6cm}p{0.6cm}}
    \toprule
        \bf Ratio & \multicolumn{1}{c}{\it 15\%} & \multicolumn{1}{c}{\it 20\%} & \multicolumn{1}{c}{\it 25\%} & \multicolumn{1}{c}{\it 30\%} & \multicolumn{1}{c}{\it 40\%} & \multicolumn{1}{c}{\it 50\%} \\
        \bf STS-B & 84.48 & 84.04 & 84.49 & \bf 84.56 & 84.48 & 83.91 \\
    \bottomrule
    \end{tabular}
    \end{center}
\vspace{-3mm}
    \caption{
        Development set results of STS-B under different masking ratio for augmentations. 
    }
    \label{tab:mask}
\vspace{-3mm}
\end{table}

\paragraph{Coefficient $\lambda$}

In Section~\ref{sec:method}, we use the $\lambda$ coefficient to weight the ELECTRA loss and then add it with contrastive loss. Because the contrastive learning objective is a relatively easier task, the scale of contrastive loss will be 100 to 1000 smaller than ELECTRA loss. As a result, we need a smaller $\lambda$ to balance these two loss terms. In the Table~\ref{tab:lambda} we show the STS-B result under different $\lambda$ values. Note that when $\lambda$ goes to zero, the model becomes a SimCSE model. We find that using $\lambda=0.005$ can give us the best performance.

\begin{table}[t]
    \begin{center}
    \centering
    \small
    \begin{tabular}{lcccccccccc}
    \toprule
        \bf $\lambda$ & \it 0 & \it 0.0001 & \it 0.0005 & \it 0.001  \\ 
        \bf STS-B & 82.22 & 83.90 & 84.40 & 84.24  \\
    \midrule
        \bf $\lambda$ & \it 0.005 & \it 0.01 & \it 0.05 & \it 0.1 \\
        \bf STS-B & \bf 84.56 & 83.44 & 84.11 & 83.66 \\
    \bottomrule
    \end{tabular}
    \end{center}
    \vspace{-3mm}
    \caption{
        Development set results of STS-B under different $\lambda$.
    }
    \label{tab:lambda}
    \vspace{-1mm}
\end{table}

\section{Analysis}
\subsection{Qualitative Study}
A very common application for sentence embeddings is the retrieval task. Here we show some retrieval examples to qualitatively explain why DiffCSE can perform better than SimCSE. In this study, we use the 2758 sentences from STS-B testing set as the corpus, and then use sentence query to retrieve the nearest neighbors in the sentence embedding space by computing cosine similarities. We show the retrieved top-3 examples in Table~\ref{tab:top-3}. The first query sentence is ``you can do it, too.''. The SimCSE model retrieves a very similar sentence but has a slightly different meaning (``you can use it, too.'') as the rank-1 answer. In contrast, DiffCSE can distinguish the tiny difference, so it retrieves the ground truth answer as the rank-1 answer. The second query sentence is ``this is not a problem''. SimCSE retrieves a sentence with opposite meaning but very similar wording, while DiffCSE can retrieve the correct answer with less similar wording.
We also provide a third example where both SimCSE and DiffCSE fail to retrieve the correct answer for a query sentence using double negation.

\begin{table}[t]
    \centering
    \scriptsize
    \begin{tabular}{p{3.5cm}|p{3.5cm}}
        \toprule
        \tf{SimCSE-BERT\ba} & \tf{DiffCSE-BERT\ba}  \\
        \midrule
        \multicolumn{2}{l}{\tf{Query}: you can do it, too.} \\
        \midrule
        1) you can use it, too. & 1) yes, you can do it. \\
        2) can you do it? & 2) you can use it, too. \\
        3) yes, you can do it. & 3) can you do it? \\
        \midrule
        \multicolumn{2}{l}{\tf{Query}: this is not a problem.} \\
        \midrule
        1) this is a big problem. & 1) i don 't see why this could be a problem. \\
        2) you have a problem. & 2) i don 't see why that should be a problem. \\
        3) i don 't see why that should be a problem. & 3) this is a big problem. \\
        \midrule
        \multicolumn{2}{l}{\tf{Query}: i think that is not a bad idea.} \\
        \midrule
        1) i do not think it's a good idea. & 1) i do not think it's a good idea . \\
        2) it's not a good idea . & 2) it is not a good idea. \\
        3) it is not a good idea . & 3) but it is not a good idea. \\
        \bottomrule
    \end{tabular}
    \vspace{-2mm}
    \caption{Retrieved top-3 examples by SimCSE and DiffCSE from STS-B test set. }
    \label{tab:top-3}
    \vspace{-3mm}
\end{table}

\begin{table}[t]
    \begin{center}
    \centering
    \small
    \begin{tabular}{l|ccc}
    \toprule
       \bf Model/Recall & \bf @1 & \bf @5 & \bf @10 \\
    \midrule
        SimCSE-BERT\ba & 77.84 & 92.78 & 95.88 \\
        DiffCSE-BERT\ba & \bf 78.87 & \bf 95.36 & \bf 97.42 \\
    \bottomrule
    \end{tabular}
    \end{center}
    \vspace{-3mm}
    \caption{
        The retrieval results for SimCSE and DiffCSE.
    }
    \label{tab:retrieval}
    \vspace{-2mm}
\end{table}

\subsection{Retrieval Task}
Besides the qualitative study, we also show the quantitative result of the retrieval task. Here we also use all the 2758 sentences in the testing set of STS-B as the corpus. There are 97 positive pairs in this corpus (with 5 out of 5 semantic similarity scores from human annotation). For each positive pair, we use one sentence to retrieve the other one, and see whether the other sentence is in the top-1/5/10 ranking. The recall@1/5/10 of the retrieval task are shown in Table~\ref{tab:retrieval}.
We can observe that DiffCSE can outperform SimCSE for recall@1/5/10, showing the effectiveness of using DiffCSE for the retrieval task.

\begin{figure}[t!]
    \centering
    \begin{subfigure}{\columnwidth}
        \includegraphics[width=\linewidth]{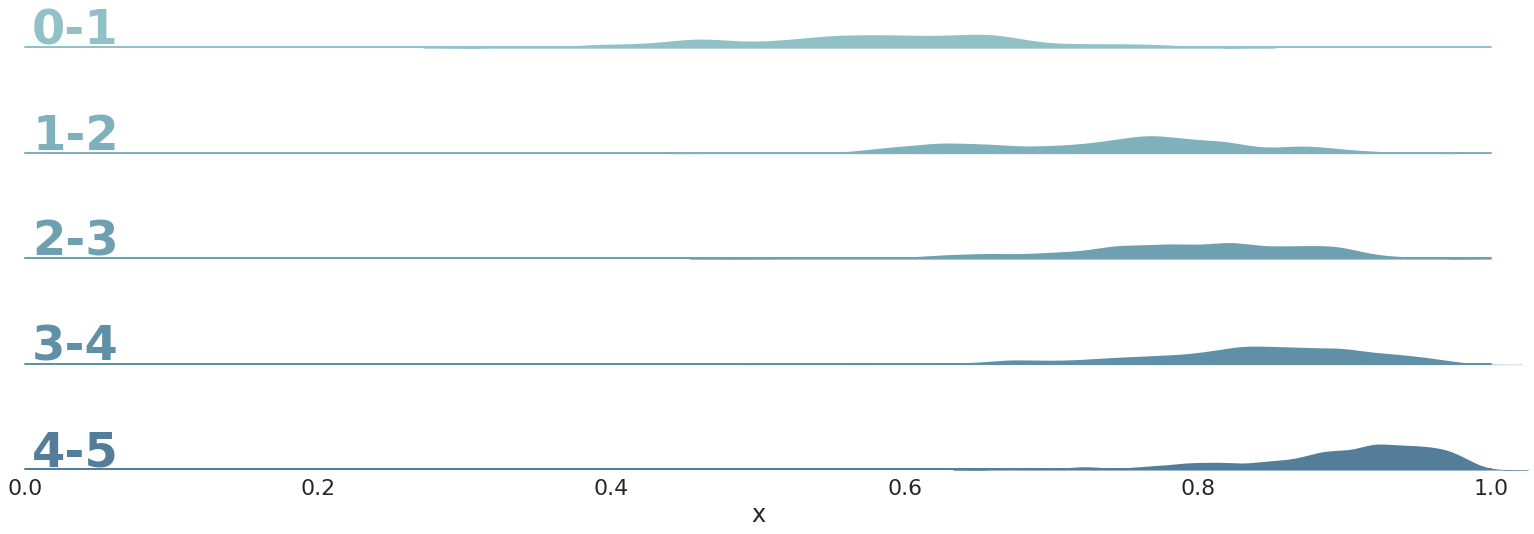}
        \caption{SimCSE}
        \label{fig:plot-dist-simcse}
        \end{subfigure}
    \begin{subfigure}{\columnwidth}
        \includegraphics[width=\linewidth]{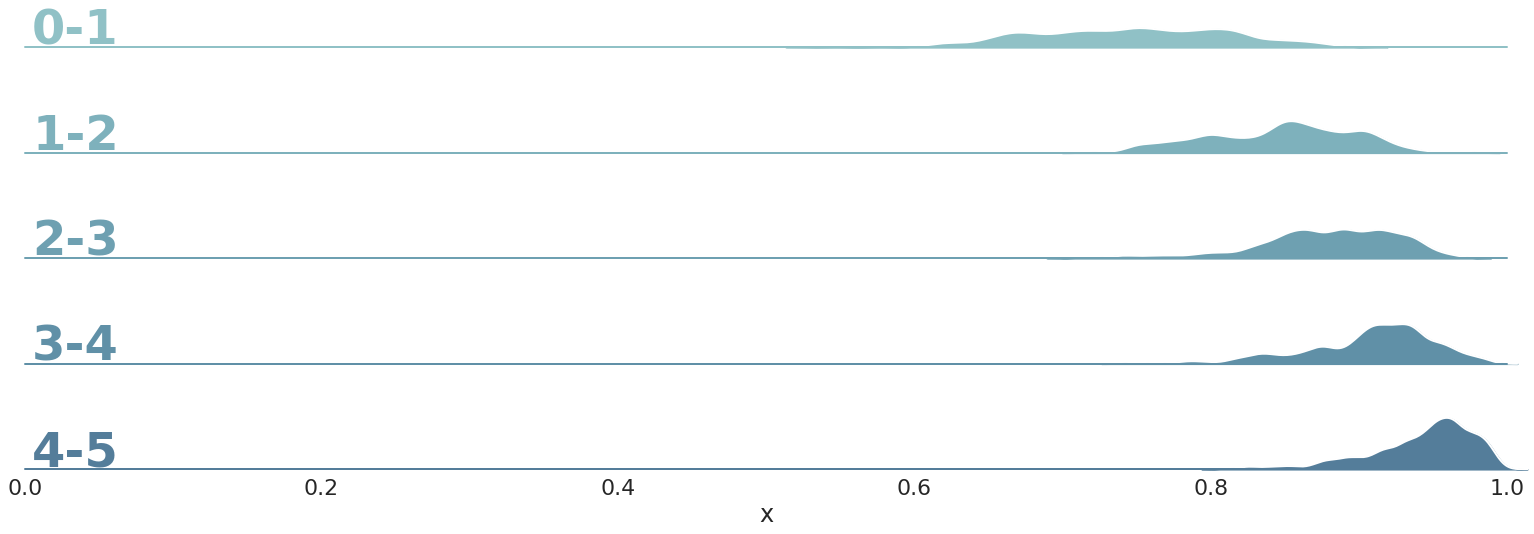}
        \caption{DiffCSE}
        \label{fig:plot-dist-diffcse}
    \end{subfigure}
    \vspace{-2mm}
    \caption{The distribution of cosine similarities from SimCSE/DiffCSE for STS-B test set. Along the y-axis are 5 groups of data splits based on human ratings. The x-axis is the cosine similarity.}
    \label{fig:plot-dist}
    \vspace{-3mm}
\end{figure}

\subsection{Distribution of Sentence Embeddings}

To look into the representation space of DiffCSE, we plot the cosine similarity distribution of sentence pairs from STS-B test set for both SimCSE and DiffCSE in Figure~\ref{fig:plot-dist}. We observe that both SimCSE and DiffCSE can assign cosine similarities consistent with human ratings. However, we also observe that under the same human rating, DiffCSE assigns slightly higher cosine similarities compared with SimCSE. This phenomenon may be caused by the fact that ELECTRA and other Transformer-based pretrained LMs have the problem of squeezing the representation space, as mentioned by~\citet{meng2021coco}. As we use the sentence embeddings as the input of ELECTRA to perform conditional ELECTRA training, the sentence embedding will be inevitably squeezed to fit the input distribution of ELECTRA. We follow prior studies~\cite{wang2020understanding, gao2021simcse} to use \emph{uniformity} and \emph{alignment} (details in Appendix~\ref{sec:uni_align}) to measure the quality of representation space for DiffCSE and SimCSE in Table~\ref{tab:uni_align}. Compared to averaged BERT embeddings, SimCSE has similar alignment (0.177 v.s. 0.172) but better uniformity (-2.313). In contrast, DiffCSE has similar uniformity as Avg. BERT (-1.438 v.s. -1.468) but much better alignment (0.097). It indicates that SimCSE and DiffCSE are optimizing the representation space in two different directions. And the improvement of DiffCSE may come from its better alignment.

\begin{table}[t]
    \begin{center}
    \centering
    \small
    \begin{tabular}{l|cc|c}
    \toprule
       \bf Model & \bf Alignment & \bf Uniformity & \bf STS \\
    \midrule
        Avg. BERT\ba & 0.172 & -1.468 & 56.70 \\
        SimCSE-BERT\ba & 0.177 & \bf -2.313 & 76.16 \\
        DiffCSE-BERT\ba & \bf 0.097 & -1.438 & \bf 78.49 \\
    \bottomrule
    \end{tabular}
    \end{center}
    \vspace{-2mm}
    \caption{
        \emph{Alignment} and \emph{Uniformity}~\cite{wang2020understanding} measured on STS-B test set for SimCSE and DiffCSE. The smaller the number is better. We also show the averaged STS score in the right-most column.
    }
    \label{tab:uni_align}
    \vspace{-3mm}
\end{table}
\section{Conclusion}

In this paper, we present DiffCSE, a new unsupervised sentence embedding framework that is aware of, but not invariant to, MLM-based word replacement. Empirical results on semantic textual similarity tasks and transfer tasks both show the effectiveness of DiffCSE compared to current state-of-the-art sentence embedding methods. We also conduct extensive ablation studies to demonstrate the different modeling choices in DiffCSE. Qualitative study and the retrieval results also show that DiffCSE can produce a better embedding space for sentence retrieval. 
One limitation of our work is that we do not explore the supervised setting that uses human-labeled NLI datasets to further boost the performance. We leave this topic for future work. 
We believe that our work can provide researchers in the NLP community a new way to utilize augmentations for natural language and thus produce better sentence embeddings.

\section*{Acknowledgements}
This research was partially supported by the Centre for Perceptual and Interactive Intelligence (CPII) Ltd under the Innovation and Technology Fund (InnoHK). 


\bibliography{anthology,custom}
\bibliographystyle{acl_natbib}

\newpage
\appendix

\section{Training Details}
\label{sec:appendix}
We use a single NVIDIA 2080Ti GPU for each experiment. The averaged running time for DiffCSE is 3-6 hours. We use grid-search of batch size $\in \{64, 128\}$ learning rate $\in \{\text{2e-6, 3e-6, 5e-6, 7e-6, 1e-5}\}$ and masking ratio $\in \{0.15, 0.20, 0.30, 0.40\}$ and $\lambda \in \{0.1, 0.05, 0.01, 0.005, 0.001\}$. The temperature $\tau$ in SimCSE is set to 0.05 for all the experiments.
During the training process, we save the checkpoint with the highest score on the STS-B development set. And then we use STS-B development set to find the best hyperparameters (listed in Table~\ref{tab:hyper_sts}) for STS task; we use the averaged score of the development sets of 7 transfer tasks to find the best hyperparameters (listed in Table~\ref{tab:hyper_trans}) for transfer tasks. All numbers in Table~\ref{tab:main_sts} and Table~\ref{tab:main_transfer} are from a single run.
\begin{table}[h!]
\centering
\small
\setlength\tabcolsep{3pt}
\begin{tabular}{ccc}
\toprule
\bf hyperparam & \bf BERT\ba & \bf RoBERTa\ba\\
\midrule
learning rate & 7e-6 & 1e-5\\
masking ratio & 0.30 & 0.20 \\
$\lambda$ & 0.005 & 0.005 \\
training epochs & 2 & 2 \\
batch size & 64 & 64 \\
\bottomrule
\end{tabular}
\caption{The main hyperparameters in STS tasks.}
\label{tab:hyper_sts}
\end{table}

\begin{table}[h!]
\centering
\small
\setlength\tabcolsep{3pt}
\begin{tabular}{ccc}
\toprule
\bf hyperparam & \bf BERT\ba & \bf RoBERTa\ba\\
\midrule
learning rate & 2e-6 & 3e-6\\
masking ratio & 0.15 & 0.15 \\
$\lambda$ & 0.05 & 0.05 \\
training epochs & 2 & 2 \\
batch size & 64 & 128 \\
\bottomrule
\end{tabular}
\caption{The main hyperparameters in transfer tasks.}
\label{tab:hyper_trans}
\end{table}

\begin{table}[h!]
\centering
\small
\setlength\tabcolsep{3pt}
\begin{tabular}{lcc}
\toprule
\bf Method & \bf BERT\ba & \bf RoBERTa\ba\\
\midrule
SimCSE & 110M & 125M\\
DiffCSE (train) & 220M & 250M\\
DiffCSE (test) & 110M & 125M\\
\bottomrule
\end{tabular}
\caption{The number of parameters used in our models.}
\label{tab:nparam}
\end{table}

During testing, we follow SimCSE to discard the MLP projector and only use the \texttt{[CLS]} output to extract the sentence embeddings.

The numbers of model parameters for BERT\ba and RoBERTa\ba are listed in Table~\ref{tab:nparam}. Note that in training time DiffCSE needs two BERT models to work together (sentence encoder + discriminator), but in testing time we only need the sentence encoder, so the model size is the same as the SimCSE model.

\paragraph{Projector with BatchNorm}
In Section~\ref{sec:ablation}, we mention that we use a projector with BatchNorm as the final layer of our model. Here we provided the PyTorch code for its structure:
\begin{lstlisting}
class ProjectionMLP(nn.Module):
    def __init__(self, hidden_size): 
        super().__init__()
        in_dim = hidden_size
        middle_dim = hidden_size * 2
        out_dim = hidden_size
        self.net = nn.Sequential(
        nn.Linear(in_dim, middle_dim, bias=False),
        nn.BatchNorm1d(middle_dim),
        nn.ReLU(inplace=True),
        nn.Linear(middle_dim, out_dim, bias=False),
        nn.BatchNorm1d(out_dim, affine=False))
\end{lstlisting}

\section{Using Augmentations as Positive/Negative Examples}

\label{sec:extrapos}
\begin{table}[t]
\centering
\small
\setlength\tabcolsep{3pt}
\begin{tabular}{lcc}

\toprule
\bf Method                 & \bf STS-B & \multicolumn{1}{c}{\bf Avg. transfer} \\
\midrule
SimCSE                     & 81.47 & 83.91                                    \\
\midrule
\it + Additional positives & \multicolumn{1}{l}{}                     \\
\midrule
MLM 15\%                   & 73.59 & 83.33                                    \\
random insert 15\%         & 80.39 & 83.92                                    \\
random delete 15\%         & 78.58 & 81.80                                    \\
\midrule
\it + Additional negatives & \multicolumn{1}{l}{}                     \\
\midrule
MLM 15\%                   & 83.02 & 84.49                                    \\
random insert 15\%         & 55.65 & 79.86                                    \\
random delete 15\%         & 55.13 & 82.56                                    \\
\midrule
\multicolumn{2}{l}{\it + Equivariance (Ours)   }                     \\
\midrule

MLM 15\%                   & \bf 84.48 & 85.95                                    \\
randomly insert 15\%         & 82.20 & 85.96                                    \\
randomly delete 15\%         & 82.59 & \bf 85.97                                    \\
\bottomrule

\end{tabular}
\caption{Development set results of STS-B and transfer tasks for using three types of augmentations (replace, insert, delete) in different ways.}
\label{tab:compare_aug}
\end{table}

In Section~\ref{sec:ablation}, we try to use different augmentations (e.g. insertion, deletion, replacement) for learning equivariance. In Table~\ref{tab:compare_aug} we provide the results of using these augmentations as additional positive or negative examples along with the SimCSE training paradigm. We can observe that using these augmentations as additional positives only decreases the performance. The only method that can improve the performance a little bit is to use MLM 15\% replaced examples as additional negative examples. Overall, none of these results can perform better than our proposed method, e.g. using these augmentations to learn equivariance.

\section{Uniformity and Alignment}
\label{sec:uni_align}

\citet{wang2020understanding} propose to use two properties, \emph{alignment} and \emph{uniformity}, to measure the quality of representations.
Given a distribution of positive pairs $p_{\mr{pos}}$ and the distribution of the whole dataset $p_{\mr{data}}$, \emph{alignment} computes the expected distance between normalized embeddings of the paired sentences:
\begin{equation*}
\ell_{\text {align }} \triangleq \underset{\left(x, x^{+}\right) \sim p_{\text {pos }}}{\mathbb{E}}\left\|f(x)-f\left(x^{+}\right)\right\|^{2}.
\end{equation*}
\emph{Uniformity} measures how well the embeddings are uniformly distributed in the representation space:
\begin{equation*}
    \ell_{\text {uniform }} \triangleq \log \underset{~~~x, y\stackrel{i.i.d.}{\sim}p_{\mathrm{data}}}{\mathbb{E}} e^{-2\|f(x)-f(y)\|^{2}}.
\end{equation*}
The smaller the values of uniformity and alignment, the better the quality of the representation space is indicated.
\section{Source Code}
\label{sec:code}

We build our model using the PyTorch implementation of SimCSE\footnote{\url{https://github.com/princeton-nlp/SimCSE}} \citet{gao2021simcse}, which is based on the HuggingFace's Transformers package.\footnote{\url{https://github.com/huggingface/transformers}} We also upload our code\footnote{\url{https://github.com/voidism/DiffCSE}} and pretrained models (links in \texttt{README.md}). Please follow the instructions in \texttt{README.md} to reproduce the results.

\section{Potential Risks}
\label{sec:risk}

On the risk side, insofar as our method utilizes pretrained language models, it may inherit and propagate some of the harmful biases present in such models. Besides that, we do not see any other potential risks in our paper.

\end{document}